\documentclass[sn-mathphys,Numbered]{sn-jnl}

\usepackage{adjustbox}
\usepackage{graphicx}
\usepackage{multirow}
\usepackage{amsmath,amssymb,amsfonts}
\usepackage{amsthm}
\usepackage{mathrsfs}
\usepackage[title]{appendix}
\usepackage{xcolor}
\usepackage{textcomp}
\usepackage{manyfoot}
\usepackage{booktabs}
\usepackage{algorithm}
\usepackage{algorithmicx}
\usepackage{algpseudocode}
\usepackage{listings}

\theoremstyle{thmstyleone}

\theoremstyle{thmstyletwo}

\theoremstyle{thmstylethree}

\begin{document}

\title[LLMs and Discourse in Engineering Design]{Opportunities for Large Language Models and Discourse in Engineering Design}

\author*[1,2]{\fnm{Jan} \sur{Göpfert}}\email{j.goepfert@fz-juelich.de}

\author[1]{\fnm{Jann M.} \sur{Weinand}}\email{j.weinand@fz-juelich.de}

\author[1]{\fnm{Patrick} \sur{Kuckertz}}\email{p.kuckertz@fz-juelich.de}

\author[1,2]{\fnm{Detlef} \sur{Stolten}}\email{d.stolten@fz-juelich.de}

\affil*[1]{\orgdiv{Institute of Energy and Climate Research -- Techno-economic Systems Analysis (IEK-3)}, \orgname{Forschungszentrum Jülich GmbH}, \orgaddress{\city{Jülich}, \postcode{52425}, \country{Germany}}}

\affil[2]{\orgdiv{Chair for Fuel Cells}, \orgname{RWTH Aachen University}, \orgaddress{Faculty of Mechanical Engineering, \city{Aachen}, \postcode{52062}, \country{Germany}}}

\abstract{
In recent years, large language models have achieved breakthroughs on a wide range of benchmarks in natural language processing and continue to increase in performance. Recently, the advances of large language models have raised interest outside the natural language processing community and could have a large impact on daily life. In this paper, we pose the question: How will large language models and other foundation models shape the future product development process? We provide the reader with an overview of the subject by summarizing both recent advances in natural language processing and the use of information technology in the engineering design process. We argue that discourse should be regarded as the core of engineering design processes, and therefore should be represented in a digital artifact. On this basis, we describe how foundation models such as large language models could contribute to the design discourse by automating parts thereof that involve creativity and reasoning, and were previously reserved for humans. We describe how simulations, experiments, topology optimizations, and other process steps can be integrated into a machine-actionable, discourse-centric design process. Finally, we outline the future research that will be necessary for the implementation of the conceptualized framework.}


\maketitle

\section{Introduction}
\label{introduction}

Large language models~(LLMs) have transformed the field of natural language processing~(NLP) and increasingly have an impact outside of academia. LLMs already dominate almost every benchmark in natural language understanding (e.g., \citep{wangSuperGLUEStickierBenchmark2019}) and current research focuses on extending them by means of other modalities such as images, videos, or sensor signals \citep{ganVisionLanguagePreTrainingBasics2022,driessPaLMEEmbodiedMultimodal2023}. Many research fields, such as medicine \citep{moorFoundationModelsGeneralist2023} and chemistry \citep{m.hockyNaturalLanguageProcessing2022}, are discussing the future implications of these models for their field. The engineering sciences are a knowledge-intensive domain, which is likely to experience great progress via the adaptation of recent methods developed in the NLP community. 
In this paper, we argue that foundation models such as LLMs can be used for creative reasoning tasks in the engineering design process, complementing and integrating existing computational methods such as topology optimization. 

\begin{figure}[H]
	\centering
	
	\includegraphics[width=0.5\linewidth]{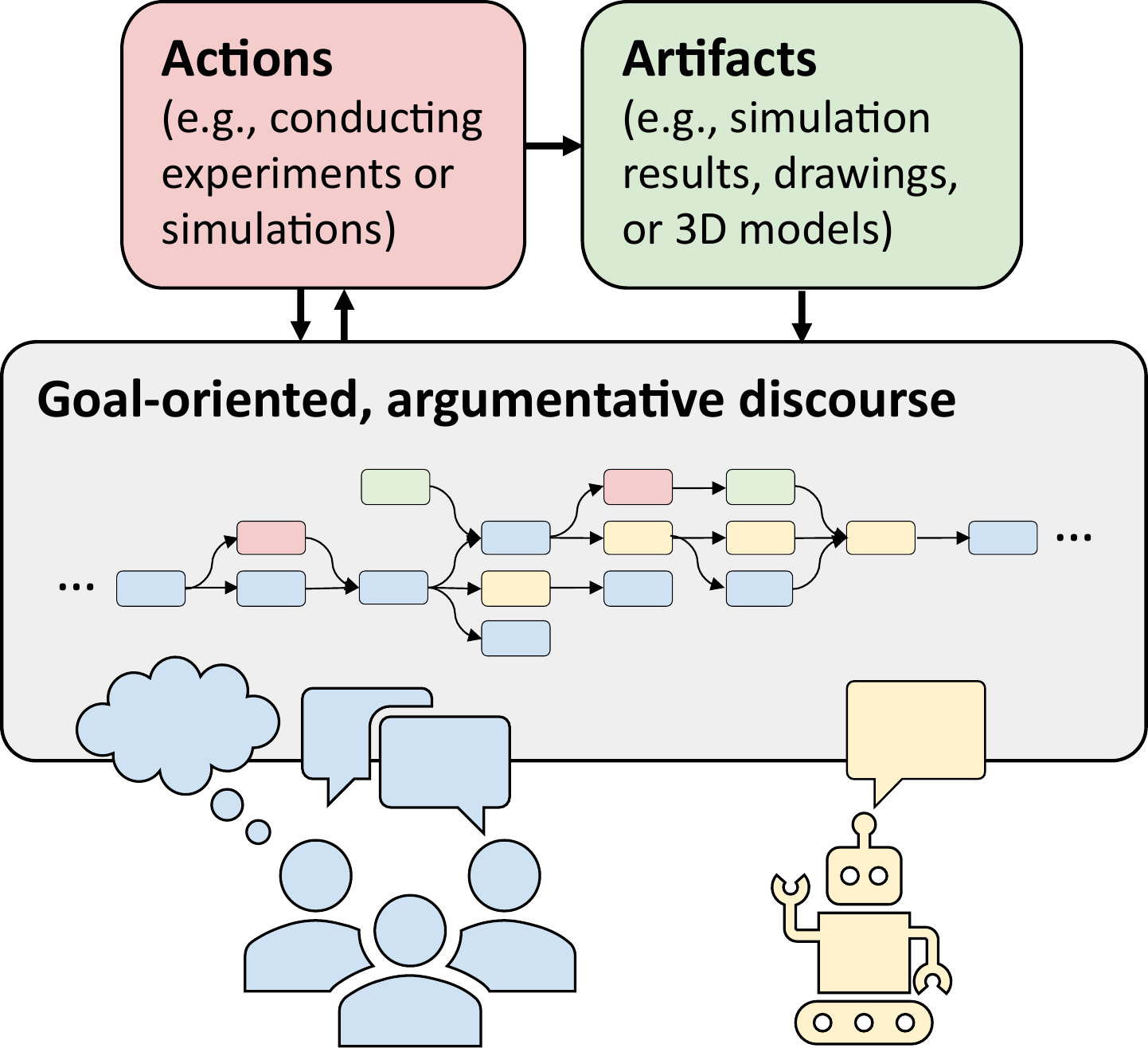}
	
	\caption[Graphical Abstract]{The engineering design process depicted as a goal-oriented, argumentative discourse formed by inter- and intrapersonal communication in which machines can participate. As part of this discourse, external actions are invoked that in turn inform it.}
	\label{fig:graphical_abstract}
\end{figure}

First, we provide engineers with a summary of the recent advances in NLP and outline which aspects of engineering design have been digitized thus far (Section~\ref{sec:background}). In Section~\ref{sec:discourse}, we place goal-oriented, argumentative discourse at the center of the product development process (see Figure~\ref{fig:graphical_abstract}) and propose making the reasoning steps explicit in the form of a new digital artifact. On this basis, we describe how LLMs and multi-modal foundation models can assist in the design discourse (Section~\ref{sec:foundation_models}) and outline interesting directions of future research (Section~\ref{sec:future_research}). The presented ideas are transferable to other contexts in which creativity and reasoning play an important role, such as scientific discovery in general.

\section{Background}
\label{sec:background}
As this article bridges multiple domains, not all of which may be familiar to most readers, we provide a thorough background for LLMs and foundation models as well as the digitization of engineering design.

\subsection{LLMs and foundation models}
In its early years, NLP applications relied on hard-coded rules. When the body of digitally-available text increased statistical methods became more prominent. Roughly from 2013 onward, NLP began to be dominated by machine-learning methods, in particular deep learning ones. \citep{manningHumanLanguageUnderstanding2022} 

Building on the distributional hypothesis, which states that words tend to have similar meanings if they occur in similar contexts \citep{harrisDistributionalStructure1954}, words (and other text units) have been represented in dense multi-dimensional vector representations using self-supervised learning \citep{mikolovEfficientEstimationWord2013,mikolovDistributedRepresentationsWords2013,penningtonGloVeGlobalVectors2014}. These word embeddings allow for the similarity between words to be calculated based on vector distances, and as they constitute a rich feature, they have been a popular component of many NLP pipelines. However, these word embeddings have been static; that is, they are context-independent (e.g., the word `seal' has the same vector representation regardless of whether it refers to the animal or machine element). Contextual word embeddings based on language models resulted in significant improvements over a wide range of NLP benchmarks \citep{petersDeepContextualizedWord2018,devlinBERTPretrainingDeep2019a}. 

In 2017, the transformer architecture was proposed \citep{vaswaniAttentionAllYou2017a}, and has been the dominant neural network architecture for language models ever since, replacing previous approaches based on feed-forward models, recurrent neural networks, or long short-term memory networks. The transformer architecture consists of an encoder and decoder. Various large language models have been proposed utilizing either the encoder (e.g., the BERT family of models \citep{devlinBERTPretrainingDeep2019a, liuRoBERTaRobustlyOptimized2019}), decoder (e.g., the GPT \citep{radfordImprovingLanguageUnderstanding,radfordLanguageModelsAre,brownLanguageModelsAre2020b}, BLOOM \citep{workshopBLOOM176BParameterOpenAccess2023} or LLaMA \citep{touvronLLaMAOpenEfficient2023} family of models), or both (e.g., the T5 and UL2 model family \citep{raffelExploringLimitsTransfer2020a,chungScalingInstructionFinetunedLanguage2022,tayUL2UnifyingLanguage2023}) of the transformer architecture.
Language models are typically trained on predicting the next token or masked tokens in a sequence in which a token can be a word, character, or a sub-word unit, with most models using sub-word-tokenization. Because models can be trained on this objective in a self-supervised setting, large unlabeled corpora (such as Wikipedia, book corpora, and Common Crawl data) can be used for training. 

Increasing the amount of data, parameters, and computation further, several emergent abilities of LLMs have been discovered \citep{weiEmergentAbilitiesLarge2022}. Sufficiently large models are capable of in-context learning \citep{brownLanguageModelsAre2020b} and chain-of-thought reasoning \citep{weiChainofThoughtPromptingElicits2023}. 
Previously requiring fine-tuning on downstream tasks (e.g., named entity recognition or question answering), LLMs can now yield decent performance on new tasks by merely including a task description and few examples in the input \citep{brownLanguageModelsAre2020b}.
Whether these abilities emerge suddenly in a sharp transition at a certain scale or in a gradual and predictable way is still subject of scientific debate \citep{schaefferAreEmergentAbilities2023}.
With instruction fine-tuning, the generated responses to a prompt aligned more with the user's intent, removing the need for careful prompt selection \citep{ouyangTrainingLanguageModels2022}. 
To solve tasks beyond the capabilities of LLMs in isolation, they have been trained to use tools for which the input and output can be represented as text, such as a calculator or Python console \citep{schickToolformerLanguageModels2023}. Further work on the agent-like behavior of LLMs combines reasoning and acting capabilities \citep{yaoReActSynergizingReasoning2023} or add self-reflection capacities \citep{shinnReflexionAutonomousAgent2023}.

The NLP community saw great advances in a wide range of benchmarks using LLMs. In addition to models operating on textual input alone, recently, multi-modal models that also process other modalities such as images, videos, and/ or sensor-signals have become a focus of research \citep{ganVisionLanguagePreTrainingBasics2022,driessPaLMEEmbodiedMultimodal2023}. Abstracting the LLM concept to other modalities, training procedures, and so forth, the term foundation model was coined \citep{bommasaniOpportunitiesRisksFoundation2022}.

\subsection{Computational engineering design}
Today, the digitization of engineering design is well advanced. In the past, technical drawing was performed on drawing boards until software for computer-aided design~(CAD) was developed in the second half of the twentieth century, which is generally adopted today. Currently, we utilize finite element analysis, topology optimization, design-support tools for additive manufacturing, and more. With model-based systems engineering and digital twins, the product development process became centered around digital models. Virtual and augmented reality enables visualization and interaction with designs. Due to the breadth of the field, only a brief overview of advances in computational engineering design can be given here, focusing on design generation, design strategy learning, and NLP (in particular LLMs) for engineering design. \\

Generative adversarial networks, feedforward neural networks, variational autoencoders, as well as reinforcement learning systems have been used in design-related generation tasks such as topology optimization or shape synthesis based on visual modalities (e.g., images, voxels, point clouds, etc.) \citep{regenwetterDeepGenerativeModels2022}.
Other work focuses on learning design strategies.
Given a state in solving a truss design problem, \citet{rainaDesignStrategyNetwork2021} predicts what actions humans perform next. \citet{gyoryHumanArtificialIntelligence2021,gyoryComparingImpactsTeam2022} analyze real time data of design teams to suggest measures from a predefined list if the communication or action frequency appears to be too low. 

Lexical databases \citep{saricaTechNetTechnologySemantic2020,jangTechWordDevelopmentTechnology2021,shiDataDrivenTextMining2017} and stopword lists \citep{saricaStopwordsTechnicalLanguage2021} for technological vocabulary and jargon have been proposed, as have engineering-related ontologies \citep{morbachOntoCAPEReUsable2009,booshehriIntroducingOpenEnergy2021,sanfilippoFormalOntologiesManufacturing2019}. With ontologies come knowledge graphs. However, there has been a lack of specialized engineering knowledge graphs thus far \citep{hanSemanticNetworksEngineering2021}. Only recently, \citet{siddharthEngineeringKnowledgeGraph2021} build a knowledge graph using patent claims. With a trend towards industry 4.0 and digital twins, the subject of the semantic representation of technological knowledge will probably be increasingly addressed in the future. 
NLP methods, which have been less applied in the engineering sciences compared to the biomedical and material ones, have become increasingly popular recently. In design research, NLP has been applied to requirements extraction, ontology construction, patent analysis, and more \citep{siddharthNaturalLanguageProcessing2022}. 

Using foundation models such as LLMs or pre-trained multi-modal models in the engineering design process is a recent and unexplored topic.
Several studies have experimented with using LLMs to provide designers with inspirational stimuli for ideation. In two explorative studies, \citet{zhuGenerativePreTrainedTransformer2022,zhuGenerativeTransformersDesign2023} prompted GPT-2 and -3 to generate design concepts~(text-to-text) based on the description of either a concept, problem, or analogy in both a fine-tuning and few-shot learning setting. Similarly, \citet{zhuBiologicallyInspiredDesign2023} fine-tune GPT-3 for bio-inspired design concept generation. \citet{maConceptualDesignGeneration2023} compare design solutions generated with GPT-3 with crowdsourced ones.
Other work has focus on design concept evaluation combining a pre-trained language model~(BERT) and image models in a multi-modal one \citep{yuanLeveragingEndUserData2021,songAttentionEnhancedMultimodalLearning2023}. 
\citet{songMultimodalMachineLearning2023} provided an extensive overview of multi-modal machine learning for engineering design. They outline possible applications, but focus on lower level tasks such as text-to-shape or shape-to-text synthesis. 

Orthogonal to the prior work, we concentrate on the design process itself as a complex, iterative, and dynamic reasoning process and situate recent advances in NLP and machine learning in a superordinate framework.

\section{Depicting the design process as a goal-oriented, argumentative discourse}
\label{sec:discourse}
We have digital artifacts for shapes, assembly processes, stress distributions, flow patterns, and more. Until now, however, computer-aided engineering, as practiced in industry, has not included the creative and argumentative process of the product development process itself. In the following, we argue that this process could be digitized and partially automatized next, and outline how this can be achieved. \\

Many steps in the product development process are performed using computation and are not based on human thought alone. However, humans are needed to integrate these computational processes, be they calculations, simulations, or optimizations, into a meaningful superordinate product development process. Human thought and world knowledge is required to reduce the solution space in advance and come up with original ideas that have not been modeled to be computationally accessible before. For example, when a bicycle is designed, the starting point is not a blank slate but an idea of how a bicycle looks and how it has worked well for over a century. If a standardized aerodynamic tube shape across bicycle manufactures is proposed, it is unlikely that this idea originated from a numerical optimization. Instead, background knowledge and the ability to think and reason are used. Solving engineering problems requires an argumentative discourse. As such, argumentation is inherent to the product development process. Experiments and calculations, etc., inform the discourse to provide necessary information. Nevertheless, argumentation is rarely given a lot of attention, perhaps because it is hidden, as it is typically not made explicit in a readable or visual representation. \\

Having described that a goal-driven, argumentative discourse is at the core of the design process, we argue that it should be represented as a digital artifact. Humans communicate, argue, and reason using natural language. Hence, the argumentative discourse can be largely represented in textual form. Many parts of the design process, however, cannot be represented as text. Therefore, we distinguish the design discourse from external actions (such as performing an experiment or simulation) and other engineering artifacts (such as a drawing, or 3D model). We formulate that external actions are invoked from within the design discourse and in turn inform the discourse, either directly or indirectly, by yielding other engineering artifacts that inform the design discourse (see Figure~\ref{fig:graphical_abstract}).

Representing the argumentative discourse as a digital artifact would improve the documentation of the design process. Instead of only archiving the results of process steps (e.g., CAD files or the results of simulation runs), the reasoning process is documented and hence archivable. For a past development process to be efficiently used for the development of a new product generation, past decisions and alternatives must be accessible. Having the reasoning process explicitly documented makes past design decisions traceable. Furthermore, making the reasoning process explicit could improve collective reasoning and therefore collaborative design. Finally, it would allow for machines to participate in the reasoning process, which the next section covers.

\section{Foundation models as interlocutors in the design discourse}
\label{sec:foundation_models}
It is apparent that LLMs and other foundation models will (and already do) reshape how we code, write, and search for information. Beyond that, we believe that they can be directly involved in the creative and argumentative design process, and so complement generative design and topology optimization tools that are already able to automatize large parts of design exploration and optimization. An example of a design discourse involving humans and machines is depicted in Figure~\ref{fig:design_discourse_example}. In the previous section, we argued that explicitly representing the argumentative discourse of the design process as a digital artifact results in many advantages. One of these is the potential for machines to participate in the reasoning process. LLMs and related multi-modal models have several characteristics that suggest that they can be successfully applied for this purpose. \\

\definecolor{human}{rgb}{0.0, 0.7, 0.0}
\definecolor{machine}{rgb}{0.0, 0.0, 1.0}
\begin{figure}
	\centering
	\footnotesize
	\definecolor{dark_gray}{rgb}{0.3,0.3,0.3}
	\framebox{\parbox{\dimexpr\linewidth-2\fboxsep-2\fboxrule}{
	\footnotesize\fontfamily{pcr}\selectfont
	\setlength{\parskip}{-5pt}
\textcolor{human}{\textbf{Prompt:}} Attached is the design of the current iteration of our aerodynamic road bike as well as the requirements, simulation results, and wind tunnel results. Please optimize the seat tube and seat stays. \\

\textcolor{machine}{\textbf{Thought:}} What to optimize for? \\
\textcolor{machine}{\textbf{Thought:}} Low aerodynamic drag has the highest priority. Other objectives are low weight, high comfort, high stiffness, low maintenance requirements. \\
\textcolor{machine}{\textbf{Action:}} Reinsuring by asking design team: Is this correct? \\

\textcolor{human}{\textbf{Prompt:}} Yes. \\

\textcolor{machine}{\textbf{Thought:}} What are the constraints? \\
\textcolor{machine}{\textbf{Action:}} Check given requirements file.\\
\textcolor{machine}{\textbf{Thought:}} The bike must comply with the rules of the governing body for sports cycling to be used in professional races.\\
\textcolor{machine}{\textbf{Action:}} Look up rule book. \\
\begin{center}
\vspace{-0.5cm}
$\vdots$ 
\vspace{0.2cm}
\end{center}
\textcolor{machine}{\textbf{Thought:}} Get inspiration from current aerodynamic road bike designs. \\
\textcolor{machine}{\textbf{Action:}} Web search \\
\textcolor{machine}{\textbf{Thought:}} Get inspiration from patents. \\
\textcolor{machine}{\textbf{Action:}} Patent search \\
\begin{center}
\vspace{-0.5cm}
$\vdots$ 
\vspace{0.2cm}
\end{center}
\textcolor{machine}{\textbf{Thought:}} How to reduce the aerodynamic drag? \\
\textcolor{machine}{\textbf{Action:}} Look up equation for aerodynamic drag. \\
\textcolor{machine}{\textbf{Thought:}} Lowering the air density, frontal area, or drag coefficient reduces aerodynamic drag. However, the air density is an environmental property which can not be changed as part of the bicycle design. \\
\textcolor{machine}{\textbf{Thought:}} The lower the seat stays the lower the frontal area. \\
\textcolor{machine}{\textbf{Thought:}} Define the lowest allowed contact point with the seat tube as a boundary condition. \\
\textcolor{machine}{\textbf{Action:}} Run multi-physics topology optimization to optimize seat stays for aerodynamics. \\
\begin{center}
\vspace{-0.5cm}
$\vdots$ 
\vspace{0.2cm}
\end{center}
\textcolor{machine}{\textbf{Thought:}} How can the drag of the seat tube be reduced? \\
\textcolor{machine}{\textbf{Thought:}} Frame and tire should flow as seamlessly as possible into each other. \\
\textcolor{machine}{\textbf{Thought:}} What problems does this solution have? \\
\textcolor{machine}{\textbf{Thought:}} High pressure zone between tire and seat tube. \\
\textcolor{machine}{\textbf{Thought:}} Is there a solution to this problem? \\
\textcolor{machine}{\textbf{Thought:}} One might drag the outer shape of the seat tube close to the tire but envelope it from behind to leave more room between the tire and the seat tube. \\
\textcolor{machine}{\textbf{Action:}} Check for patent infringements \\

\textcolor{human}{\textbf{Prompt:}} Can you visualize this idea? \\
\begin{center}
\vspace{-0.5cm}
$\vdots$ 
\vspace{0.2cm}
\end{center}
}}
$\text{\adjustbox{bgcolor=human}{\phantom{text}}}=\text{Human}$\quad $\text{\adjustbox{bgcolor=machine}{\phantom{text}}}=\text{Machine}$
	\caption{An example of a design discourse. Note that the system outputs should clarify the ideas presented here and are not real outputs.}
	\label{fig:design_discourse_example}
\end{figure}

LLMs and related multi-modal models can input and output natural language. This facilitates their integration into the argumentative discourse, which to a large extent is conducted in natural language anyway, be it in intra- or interpersonal communication. World knowledge is important both to interact with humans, where a certain common ground is required, and for reducing the solution space and coming up with creative solutions. Preliminary evidence suggests that LLMs possess rich representations of the world despite being trained on simple objectives \citep{liEmergentWorldRepresentations2023}. It is evident that humans are constrained by their knowledge when designing by analogy or biomimicry. In contrast, LLMs can accumulate wide-ranging knowledge during training. 
LLMs of sufficient size display strong results on various tasks involving reasoning and are able to perform step-by-step reasoning \citep{weiChainofThoughtPromptingElicits2023}. 
Multi-modal models can operate on various forms of design representations, which is important, because throughout the product development process, designers utilize different types of representations of their designs (i.a., text, tables, sketches, and 3D models) \citep{songMultimodalMachineLearning2023}.
Finally, many tasks in the engineering design process cannot be solved by means of pure thought but require specialized engineering software and databases (i.a., CAD and simulation software, patent and material databases). Recent work shows that LLMs can be trained to interact with APIs (including the decision of when to call which API with which arguments) in a self-supervised setting \citep{schickToolformerLanguageModels2023}. \\

We conclude that these models are fundamentally applicable to the purpose of assisting in the engineering design discourse. Nevertheless, a single call to a model will not be of great utility; instead, the models must be embedded within a framework to solve complex engineering tasks. In the next section, we outline aspects of such a framework and simultaneously highlight promising future research directions.

\section{Recommendations for future research}
\label{sec:future_research}
Many puzzle pieces for the outlined transformation of the engineering design process are at hand. However, several aspects, which have not been sufficiently researched, are likely to be necessary for a successful implementation of the proposed concept, namely:

\paragraph{Formalizing the engineering design discourse within a framework of stages and components.}
Although approaches to interpret how outputs are formed in deep neural network models (such as probing \citep{belinkovProbingClassifiersPromises2022}) exist, in practice, LLMs largely constitute black boxes. To increase the interpretability and accuracy of LLMs, approaches such as scratchpad \citep{nyeShowYourWork2021} or chain-of-thought prompting \citep{weiChainofThoughtPromptingElicits2023} steer the models towards the generation of intermediate steps. 
Complementary to the aforementioned approaches of generating intermediate steps, models can be guided to imitate certain patterns of thinking or to follow a logical flow by embedding calls to LLMs into a framework with a predefined causal structure. In such frameworks ``querying a language model becomes a computational primitive.'' \citep{creswellFaithfulReasoningUsing2022} 
For example, answering questions in steps adhering to formal logic yields interpretable reasoning traces and reduces the ``hallucination'' of facts \citep{creswellFaithfulReasoningUsing2022}.
Similarly, formalizing the engineering design discourse within a framework of stages and recurring components will help safeguard the models and increases the detail at which reasoning processes can be documented and verified. Furthermore, it contributes to the clarity of the discourse and makes LLMs more controllable, in that humans have more points of intervention, allowing them to provide more fine-grained guidance. For defining such a framework, the extensive literature on design research can assist (e.g., \citep{frickeSuccessfulIndividualApproaches1996,pahlEngineeringDesign2007}).

\paragraph{Creating machine-actionable interfaces for engineering software.} 
We noted that using external tools via APIs with LLMs is already an active area of research. The tools invoked in the design discourse could include specialized engineering software such as topology optimization. However, for today's LLMs, the tools are required to provide a textual interface; that is, both inputs and outputs are represented as text \citep{schickToolformerLanguageModels2023}. Therefore, the interfaces of specialized engineering software should be adapted according to the models' requirements. Note that the requirements for an interface are bound to change with multi-modal models.

\paragraph{Learning representations better corresponding to skills required for engineering.}
Good spatial imagination, solid engineering knowledge, and high heuristic competence are skills that correlate with higher quality engineering design solutions \citep{frickeSuccessfulIndividualApproaches1996}. For foundation models to fulfill a supportive role in the engineering design process, the learned representations should in turn correlate to a high degree with these skills. Therefore, the proportion of technical documents (such as patents or specialist books) in the pre-training corpus can be increased. However, although LLMs compress large amounts of knowledge in their weights, a representation corresponding to good spatial imagination and causal understanding of physical processes is unlikely to be learned from text alone. With a trend towards multi-modal models, incorporating images, videos, and other modalities, we see this problem as likely to be addressed in the near future. Additionally, strengthening the reasoning capabilities of LLMs is a promising and growing research direction.

\paragraph{Specifying evaluation metrics and datasets.}
Prior work has expressed the need for design-specific metrics for evaluating deep generative models in design synthesis \citep{zhuGenerativePreTrainedTransformer2022,songMultimodalMachineLearning2023,regenwetterStatisticalSimilarityRethinking2023, maConceptualDesignGeneration2023}. Similarly, design-specific metrics are likely to be required to evaluate the performance of frameworks to assist in the argumentative design discourse. In cases where ground truth is not required, more complex assessments, for which there is not yet a calculable metric, but for which human evaluators are required, such as the assessment of usefulness and feasibility \citep{maConceptualDesignGeneration2023}, could possibly be approximated by foundation models in the future. 
Furthermore, as participation in the argumentative discourse of engineering design is a new task, there is a need for public datasets to evaluate and compare models against.

\section{Conclusion}
\label{sec:conclusion}
We propose conceiving the design process as a goal-oriented, argumentative discourse on which foundation models can operate. Making the reasoning steps explicit in a new digital artifact of the product development process could lead to improved documentation and increased collaboration, and makes certain forms of machine assistance possible in the first place. We describe how LLMs and multi-modal foundation models can assist in the design discourse and outline interesting directions of research, including structuring of the design discourse in stages and components, the provision of machine-actionable interfaces for specialized engineering software, the development of foundation models with learned representations that better correspond to the skills required in the design process, and the creation and publication of common metrics and datasets as a community effort. We do not intend to describe a theoretical framework that is far from ever being implemented in practice, but believe that the product development process is about to change, with the aspects described in this article being part of its future.

\backmatter

\bmhead{Acknowledgments}

The authors would like to thank the German Federal Government, the German state governments, and the Joint Science Conference~(GWK) for their funding and support as part of the NFDI4Ing consortium. Funded by the German Research Foundation~(DFG) -- project number: 442146713. Furthermore, this work was supported by the Helmholtz Association under the program “Energy System Design”.

\begin{appendices}

\end{appendices}

\bibliography{llms_for_engineering_design_perspective} 

\end{document}